\documentclass{article}

\usepackage[preprint,nonatbib]{neurips_2023}

\usepackage[sort,numbers]{natbib}
\usepackage{amssymb}
\usepackage[utf8]{inputenc} % allow utf-8 input
\usepackage[T1]{fontenc}    % use 8-bit T1 fonts
\usepackage{hyperref}       % hyperlinks
\usepackage{url}            % simple URL typesetting
\usepackage{booktabs}       % professional-quality tables
\usepackage{amsfonts}       % blackboard math symbols
\usepackage{nicefrac}       % compact symbols for 1/2, etc.
\usepackage{microtype}      % microtypography
\usepackage[table]{xcolor}
\usepackage{tikz}
\usepackage{xcolor}         % colors
\usepackage{tabularx}
\usepackage{makecell}
\usepackage{natbib}

\usepackage{bm}
\usepackage{url}            % simple URL typesetting
\usepackage{amsfonts,amsmath}       % blackboard math symbols
\usepackage{nicefrac}       % compact symbols for 1/2, etc.
\usepackage{microtype}      % microtypography
\usepackage{xcolor}
\usepackage{tabu}
\usepackage{multirow}       % multiple rows in table
\usepackage{graphicx}
\usepackage{graphics}
\usepackage{tabularx}
\usepackage{makecell}
\usepackage{caption}
\usepackage{wrapfig}
\usepackage{enumitem}
\usepackage{subcaption}
\usepackage{arydshln}
\usepackage{algpseudocode}
\usepackage{color, colortbl}

\usepackage{listings}
\definecolor{Gray}{gray}{0.95}
\definecolor{orange}{rgb}{0.9,0.5,0}

\newcommand{\beq}{\vspace{0mm}\begin{equation}}
\newcommand{\eeq}{\vspace{0mm}\end{equation}}
\newcommand{\beqs}{\vspace{0mm}\begin{eqnarray}}
\newcommand{\eeqs}{\vspace{0mm}\end{eqnarray}}
\newcommand{\barr}{\begin{array}}
\newcommand{\earr}{\end{array}}

\usepackage{color, colortbl}
\definecolor{Gray}{gray}{0.93}

\usepackage{xcolor,amsmath}
\usepackage[linesnumbered,ruled,vlined]{algorithm2e}
\DontPrintSemicolon

\SetKwComment{Comment}{\color{green!50!black}\# }{}

\SetKwProg{Function}{def}{:}{}

\SetKwProg{For}{for}{:}{}
\SetKwProg{If}{if}{:}{}

\usepackage{pifont}% http://ctan.org/pkg/pifont
\title{Efficient Mixed Transformer for Single Image Super-Resolution}

\author{%
	Ling Zheng$^1$, Jinchen Zhu$^{1,2}$, Jinpeng Shi$^{1,2}$, Shizhuang Weng$^1$\\
	\small{$^1$Anhui University, $^2$Fried Rice Lab} \\
	\texttt{weng\_1989@126.com} \\
	\begin{small}
	\url{https://jinchen2028.github.io/EMT/}
    \end{small}
}

\begin{document}

\maketitle

\begin{abstract}
Recently, Transformer-based methods have achieved impressive results in single image super-resolution (SISR). However, the lack of locality mechanism and high complexity limit their application in the field of super-resolution (SR). To solve these problems, we propose a new method, Efficient Mixed Transformer (EMT) in this study. Specifically, we propose the Mixed Transformer Block (MTB), consisting of multiple consecutive transformer layers, in some of which the Pixel Mixer (PM) is used to replace the Self-Attention (SA). PM can enhance the local knowledge aggregation with pixel shifting operations. At the same time, no additional complexity is introduced as PM has no parameters and floating-point operations. Moreover, we employ striped window for SA (SWSA) to gain an efficient global dependency modelling by utilizing image anisotropy. Experimental results show that EMT outperforms the existing methods on benchmark dataset and achieved state-of-the-art performance.  
\end{abstract}

\section{Introduction}
\begin{wrapfigure}{r}{0.45\textwidth}
	\vspace{2mm}
	\begin{center}
		\vspace{-7.5mm}
		\hspace{-4mm}
		\includegraphics[width=0.44\textwidth]{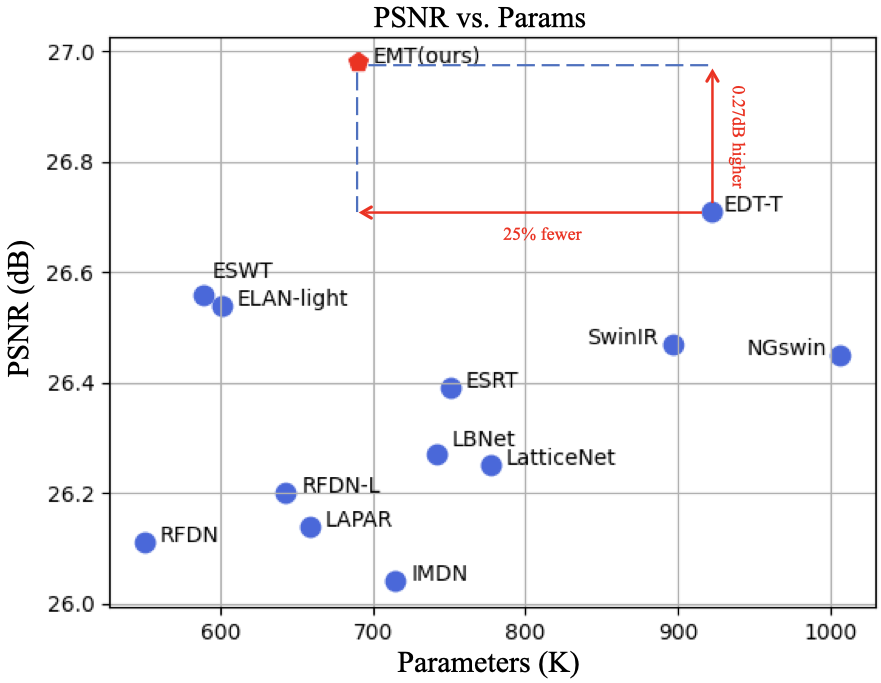}
	\end{center}
	\vspace{-3mm}
	\captionsetup{font=footnotesize}    
	\caption{Comparison of the trade-off between model performance and complexity on the Urban100 \cite{U100} ($\times4$) test set.}
	\vspace{-5mm}
	\label{fig_1compare}
\end{wrapfigure}
The purpose of Single Image Super-Resolution (SISR) is to recover high-resolution (HR) from low-resolution (LR) images \cite{SRphoto}.
Convolutional neural network (CNN) \cite{DeepSR0,DeepSR1,DeepSR2,DeepSR3,DeepSR4,VDSR,SRusedconv1,SRusedconv2} -based Super-Resolution (SR) methods are popular because of their powerful ability to extract high frequency detail from images. However, establishing global connectivity by using CNN-based methods \cite{SRCNN,EDSR,VDSR} is difficult. 
As an alternative, Transformer-based methods \cite{transformer,swinir,elan,EDSR} exploit powerful Self-Attention (SA) to model global dependencies on the input data, and has shown impressive performance.

Recently, several studies found that the Transformer lacks  locality mechanism for information aggregation within local regions \cite{Local_SA1,Local_SA2}. Local knowledge is highly relevant to the structure and details of the image and crucial for SISR. As more and more SR networks are required to be loaded onto mobile or embedded devices, lightweight SR (LSR) methods have gradually become research hotspots. Many transformer-based lightweight networks, such as ESRT \cite{ESRT}, SwinIR \cite{swinir}, and ELAN-light \cite{elan} have been proposed.
These methods reduce the complexity by modifying the SA calculation, such as using a non-fixed computational window or shared attention mechanism for multiple SAs. However, the modified networks remain highly complex, and SA is an expensive module for LSR application.

The above discussion leads to a significant research hotspot regarding Transformer-based LSR methods: how to enhance the necessary locality mechanism and gain efficient global dependency modelling while reducing complexity. To solve this problem, we propose the Efficient Mixed Transformer (EMT) for SISR. First, we propose the Mixed Transformer Block (MTB) with multiple consecutive transformer layers, where the SAs in several layers are replaced with local perceptrons to improve the overall local knowledge aggregation. Second, we develop a Pixel Mixer (PM) using channel segmentation and pixel shifting as the local perceptron. PM expands the local receptive field by fusing adjacent pixel knowledge from different channels to improve the locality mechanism. Notably, PM reduced the complexity of the overall network given its lack of additional parameters and floating-point operations (FLOPs). Third, we exploit striped window for SA (SWSA) by using the anisotropic feature of the image to improve the efficiency of global dependency modelling.  Finally, extensive experiments show that our method achieves better performance with fewer parameters than the existing efficient SR methods, as shown in Fig. \ref{fig_1compare}.

\section{Related Work}
\paragraph{Locality Mechanism in Transformers.}
Previous studies have shown that capturing local spatial knowledge using transformer-based methods is difficult, limiting their application in the field of SR. Several attempts have been made to introduce locality in the Transformer-based networks \cite{DWconv_SA, Local_SA1, Local_SA2}. \citet{Local_SA1} bring in depth-wise convolution in feed-forward network to improve the overall locality and achieve competitive results in ImageNet classification \cite{imagenet}. Later, \citet{DWconv_SA} propose to replace the SA in the Swin Transformer with a depth-wise convolution and achieve comparable performance in high-level computer vision tasks to Swin Transformer \cite{swintrans}. Inspired by these works, we explore local perceptrons that enhance the local knowledge aggregation of the network, such as convolution, to replace the SA and thus improve locality mechanism.

\paragraph{Transformer-based method for LSR.}
Recently, Transformer-based LSR methods have been proposed. \citet{swinir} applies the Swin Transformer \cite{swintrans} structure to LSR and propose SwinIR, achieving impressive results by exploiting window-based attention mechanisms. \citet{ESRT} developed Efficient Multi-Head Attention (EMHA) to reduce the use of training data and lower the memory occupation of the GPU. Then, \citet{elan} proposes group-wise multi-scale self-attention (GMSA) by using different window sizes and shared attention mechanisms to solve the redundancy of SA computation. However, the modified transformer-based methods are still complex.
%-------------------------------------------------------------------------

\section{ Methodology}
\label{method}
\begin{figure*}[t]
	\centering
	\begin{subfigure}{1\linewidth}
		\centering
		\includegraphics[width=1\linewidth]{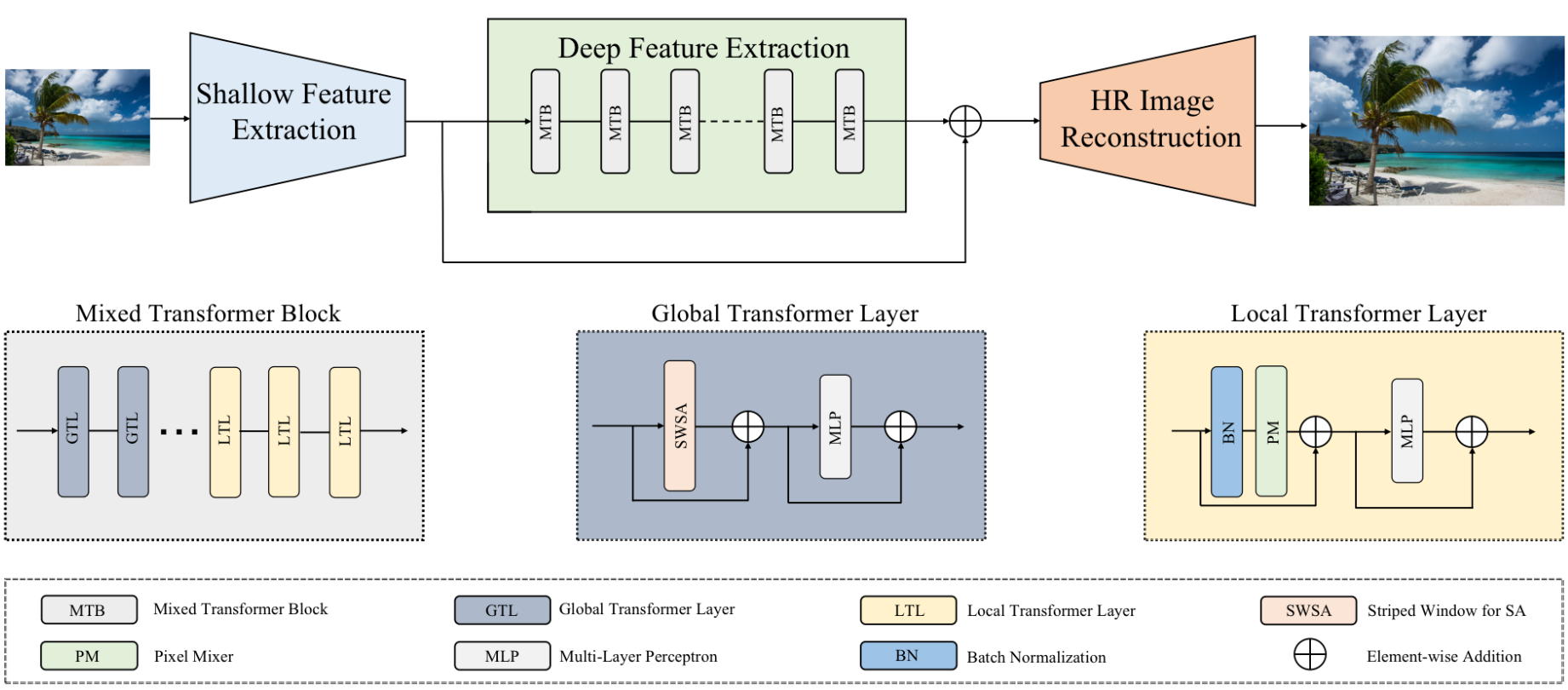}
		\label{zhengti}%文中引用该图片代号
	\end{subfigure}
	\caption{EMT architecture for image SR.}
	\label{EMT}
\end{figure*}
\subsection{Overall EMT Architecture}
As shown in Fig. \ref{EMT}, EMT consists of three parts: shallow feature extraction unit (SFEU), deep feature extraction unit (DFEU), and reconstruction unit (RECU). We use $I_{lr}\in\mathbb{R}^{H \times W \times C_{in}}$ and  $I_{sr}\in\mathbb{R}^{H \times W \times C_{in}}$  as the input and output of EMT, respectively. We only use a $3\times3$ convolutional layer as a SFEU to process the input image $F_0\in\mathbb{R}^{H \times W \times C}$: 
\begin{equation}
	F_0=H_{SF}(I_{lr}),
\end{equation}
where $H_{SF}(\cdot)$ denotes the function of SFEU; $H$, $W$, and $C_{in}$ denotes the height, width, and number for channels of the input LR image; $C$ denotes the number of channels of intermediate features.

Subsequently, $F_0\in\mathbb{R}^{H \times W \times C}$  is extracted by DFEU to obtain the deep features $F_D\in\mathbb{R}^{H \times W \times C}$, and the unit contains $n$ MTB. The processing formula is as: 
\begin{equation}
	\begin{split}
		{F_{D}}& =H_{DF}(F_0),\\
		&=H_{MT{B_n}}(H_{MT{B_{n-1}}}({\cdot\cdot\cdot}H_{MT{B_1}}(F_0){\cdot\cdot\cdot}));
	\end{split}
\end{equation}
where $H_{DF}(\cdot)$ is the DFEU function and $H_{MT{B_n}}(\cdot)$ represents the $n$-th MTB in the DFEU. MTB consists of multiple consecutive transformer layers, where SA is replaced in some of the layers.\\
Finally, as $F_0$ and $F_D$ are rich in low and high frequency information, they are summed and transmitted directly to RECU:
\begin{equation}
	I_{sr}=H_{REC}(F_0+F_D),
\end{equation}
where $H_{REC}$ represents the processing function of the RECU.

The EMT is then optimized using a loss function, where many loss functions are available, such as $L_2$ \cite{SRusedconv1,SRusedconv3,MemNet,SRusedconv2}, $L_1$ \cite{SRusedconv4,EDSR,RDN,elan,swinir}, and perceptual losses \cite{PLoss1,PLoss2}. For simplicity and directness, we select the $L_1$ loss function. Given a training set $\{I_{LR}^i,I_{HR}^i\}_{i=1}^N$ with a total of N ground-truth HR and matching LR images, the parameters of EMT are trained by minimizing the $L_1$ loss function:
\begin{equation}
	\mathcal{L}= \frac{1}{N}\sum_{i=1}^{N}\|I_{RHR}^i-I_{HR}^i\|_1,
\end{equation}
where $I_{RHR}$ is the EMT output of $I_{LR}$.
\begin{algorithm}[b]
	\caption{Pixel Mixer for EMT, PyTorch-like Code}
	\label{PM-alg}
	\definecolor{codeblue}{rgb}{0.25,0.5,0.5}
	\definecolor{codekw}{rgb}{0.85, 0.18, 0.50}
	\lstset{%
		backgroundcolor=\color{white},
		basicstyle=\fontsize{7.5pt}{7.5pt}\ttfamily\selectfont,
		columns=fullflexible,
		breaklines=true,
		captionpos=b,
		commentstyle=\fontsize{7.5pt}{7.5pt}\color{codeblue},
		keywordstyle=\fontsize{7.5pt}{7.5pt}\color{codekw},
	}
	\begin{lstlisting}[language=python]
import torch
		
class PixelMixer(torch.nn.Module):
	def __init__(self):
		super().__init__()
		# list of shift rules
		self.rule =  [[-1, 0], [0, 1], [0, -1],
	        [1, 0], [0, 0]]
		
	def forward(self, x):
		groups = torch.split(x, [x.shape[1]//5] * 5, dim=1)
		# use different shift rules for each group
		groups = [torch.roll(group, shifts=rule, dims=(2, 3))
		for group, rule in zip(groups, self.rule)]
		return torch.cat(groups, dim=1)
	\end{lstlisting}
\end{algorithm}
\subsection{Mixed Transformer Block for SR}
 Starting from \cite{swintrans}, many works have optimized SA and achieved good results in various computer vision tasks. However, the modified SA still cannot address the lack of locality mechanism in Transformer-based methods and still remains high complexity. Thus, we propose Mixed Transformer Block (MTB), which consists of two types of transformer layers, namely the Local Transformer Layer (LTL) and Global Transformer Layer (GTL). In LTL, we use local perceptrons to replace SA, thereby improving overall local knowledge aggregation and reducing the complexity of layers. In addition, a new local perceptron, PM, with no computational cost is developed. For GTL, we use striped window for SA to efficiently build global dependency modelling.

\subsection{Pixel Mixer}
\citet{Shiftc} introduces locality in the network by proposing shift convolution  instead of spatial convolution, achieving competitive performance in high-level computer vision tasks. On the base of this work, we extended the idea and developed PM by improving it. Specifically, PM first divides the feature channels into five equal groups, then shifts the feature points of the first four groups in a specific order (left, right, top, bottom) and fill the blank pixels on the opposite side with those that are out of range. By exchanging several channels between adjacent features, the surrounding knowledge is mixed and the channel blending module is expanded with the receptive field to quickly capture local spatial knowledge. In addition, by associating edge feature points with the opposite ones each input window in the self-attention mechanism can obtain different knowledge from other source.

We assume that $z\in{H \times W \times C}$, where $H$, $W$ and $C$ represent the height, width and number of channels, respectively; and $z'$ represents the output with the same shape as the input. The equation is as follows:
%\begin{equation}
%	\begin{split}
%	\small
	%z[[0 : H - n] : [H - n : H], 0 : W, 0 : \beta C]\ \ \ \ \ &\rightarrow z'[[H - n : H] : [0 : H - n], 0 : W, 3\beta C : 4\beta C] \\
	%z[0 : H, [ 0 : W - n ] : [ W - n : W], \beta C : 2\beta ] &\rightarrow z'[0 : H, [ W - n : W] : [ 0 : W - n ], 0 : \beta C] \\
	%z[0 : H, [ 0 : n ] : [ n :W], 2\beta C : 3\beta C]\ \ \ \ \ \ \ \ \ \ \ \ \ &\rightarrow z'[0 : H,[ n :W] : [ 0 : n ], \beta C : 2\beta C]\\
	%z[[0 : n] : [n : H], 0 : W, 3\beta C : 4\beta C]\ \ \ \ \ \  \ \ \ \ \ \ \ &\rightarrow z'[[n : H] : [0 : n], 0 : W, 2\beta C : 3\beta C]\\
	%z[0 : H, 0 : W, 4\beta C : C] \ \ \ \ \ \ \ \ \ \ \ \ \ \ \ \ \ \ \  \ \ \ \ \ \ \ \ \ \ \  \ \ \ \ &\rightarrow z'[0 : H, 0 : W, 4\beta C : C],
   % \end{split}
%\end{equation}
\begin{small} 
\begin{equation}
	\begin{aligned}
		&z[[0 : H - n] : [H - n : H], 0 : W, 0 : \beta C]&&\rightarrow z'[[H - n : H] : [0 : H - n], 0 : W, 3\beta C : 4\beta C] \\
		&z[0 : H, [ 0 : W - n ] : [ W - n : W], \beta C : 2\beta ]&&\rightarrow z'[0 : H, [ W - n : W] : [ 0 : W - n ], 0 : \beta C]\! \\ 
		&z[0 : H, [ 0 : n ] : [ n :W], 2\beta C : 3\beta C]&&\rightarrow z'[0 : H,[ n :W] : [ 0 : n ], \beta C : 2\beta C]\\
		&z[[0 : n] : [n : H], 0 : W, 3\beta C : 4\beta C]&&\rightarrow z'[[n : H] : [0 : n], 0 : W, 2\beta C : 3\beta C]\\
		&z[0 : H, 0 : W, 4\beta C : C]&&\rightarrow z'[0 : H, 0 : W, 4\beta C : C],
	\end{aligned}
\end{equation}
\end{small}

where $n$ represents the shift step, set to $1$ pixel in this paper, and $\beta$ is the parameter that controls the percentage of channel shifted, in this case $\frac{1}{5}$.
 
This process is outlined in Algorithm \ref{PM-alg}. PM is simple and clean, with no additional parameters and FLOPs during operation, and it is thus a practical solution for effectively improving SR network performance.

\begin{figure*}[t]
	\centering
	\begin{subfigure}{1\linewidth}
		\centering
		\includegraphics[width=1\linewidth]{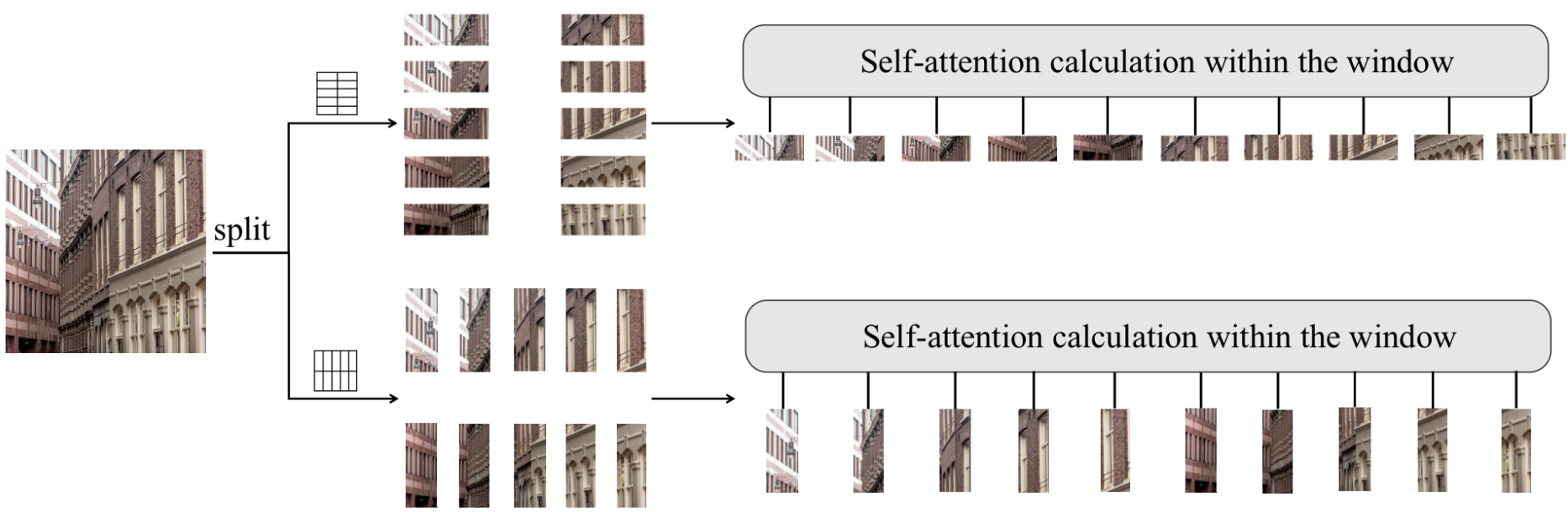}
	\end{subfigure}
	\caption{SWSA can divide the window by taking advantage of the anisotropic image features. For simplicity, only one case of the striped window is shown here.}
	\label{striped_SA}%文中引用该图片代号
\end{figure*}

\subsection{Striped Window for SA}
Transformer is particularly good at global dependency modelling. Although the global connectivity between token embeddings can be computed by self-attention mechanism, anisotropic features in the image still make the isotropic square window SA redundant \cite{IR-stripes-SA}. To efficiently model global dependencies, we propose to use mutually perpendicular striped window for SA (SWSA). Specifically, the span of multi-scale similarity or symmetry in an image is horizontally or vertically anisotropic, so it is difficult to fully utilize this feature using horizontal or vertical windows \cite{ESWT} alone. we use mutually perpendicular windows and multi-head calculations within each window to cope with this problem. As shown in Fig. \ref{striped_SA}, the striped window follows the anisotropic feature of the image, making the proportion of similar features within the window larger, and the computed attention score increases its focus on modeling the contained features. In addition, The computational efficiency is further improved by applying multi-head calculations within each window.

For an input image size $H \times W \times C$, the $H$, $W$ and $C$ represent the height, width, and number of channels, respectively. The self-attention of each window is then calculated separately to obtain the output $F_{out_n}\in\mathbb{R}^{C \times H \times W}$, where $n$ represents the $n$th of the 2 striped windows. To calculate the query and value matrices $Q$ and $V$, the following steps are performed: 
Reshape the input $X\in\mathbb{R}^{C\times H \times W}$ into $X'_n\in\mathbb{R}^{C/2\times (H\times W)}$.
Calculate the linear transformation of $X'$ using weight matrices $W_Q$ and $W_V$:
\begin{equation}
	Q_n = X'_n \times W_Q , V_n = X'_n \times W_V ,
\end{equation}
where $Q_n \in \mathbb{R}^{d_q\times (H\times W)}$ and $V_n\in\mathbb{R}^{d_v\times (H\times W)}$, respectively. Then multi-head calculations are applied to each window.

To calculate the self-attention score matrix $F_{out_n}$, the softmax function was applied as follows:
\begin{equation}
F_{out_{n}} =  Softmax( \frac{Q_{n} \cdot Q_{n}^T}{scale})V_{n},
\end{equation}
where $scale$ is a constant used to control the size of the matrix $A$.
The $F_{out_{n}}$ matrix is concatenated along the channel dimension to obtain the output of SWSA:

\begin{equation}
F_{out} = [F_{out_1}; F_{out_n} ],
\end{equation}
note that the $n$ windows are computed sequentially, and the final results are concatenated. 
\begin{table*}[t]
	\scriptsize
	\centering
	\caption{Quantitative comparison with SOTA LSR methods on benchmark datasets of $\times3$ and $\times4$. The best results are marked in {\color{red}red} and the second-best ones are in {\color{blue}{blue}}.}
	\label{tab: comparison}
	\resizebox{1\linewidth}{!}{
		\begin{tabu}{l|c|c|cc|cc|cc|cc|cc}
			\hline
			\specialrule{0em}{2pt}{0pt}
			
			\multirow{2}{*}{Method}     & \multirow{2}{*}{Scale}    & Params  & \multicolumn{2}{c|}{Set5}  & \multicolumn{2}{c|}{Set14} & \multicolumn{2}{c|}{BSD100} & \multicolumn{2}{c|}{Urban100} & \multicolumn{2}{c}{Manga109} \\
			&                             &(K)                        & PSNR                      & SSIM                       & PSNR                        & SSIM                          & PSNR                            & SSIM            & PSNR           & SSIM            & PSNR           & SSIM \\
			
			\specialrule{0em}{0pt}{0pt}
			\hline
			\specialrule{0em}{1pt}{0pt}
			\hline
			\specialrule{0em}{2pt}{0pt}

			FDIWN \cite{FDIWN}           & \multirow{9}{*}{$\times3$}                            & 645                  & 34.52                     & 0.9281                     & 30.42                       & 0.8438                        & 29.14                           & 0.8065          & 28.36          & 0.8567          & -              & -  \\
			LBNet \cite{LBNet}           &                             & 736                      & 34.47                     & 0.9277                     & 30.38                       & 0.8417                        & 29.13                           & 0.8061          & 28.42          & 0.8559          & 33.82          & 0.9460  \\
			LAPAR-A \cite{LAPAR}           &                             & 544                  & 34.36                     & 0.9267                     & 30.34                       & 0.8421                        & 29.11                         & 0.8054         & 28.15          & 0.8523          & 33.51         & 0.9441  \\
			NGswin \cite{NGswin}                &                             &  1,007                  &  34.52            & 0.9282             &  30.53                &  0.8456                &  29.19             &  0.8078             &  28.52              &  0.8603                             &  33.89  &  0.9470   
			\\
			SwinIR \cite{swinir}         &                             & 886                     & 34.62                     & 0.9289                     & 30.54                       & 0.8463                        & 29.20                           & 0.8082          & 28.66          & 0.8624          & 33.98          & 0.9478  \\
			ELAN-light \cite{elan}       &                             & 590                & 34.61                     & 0.9288                     & \color{blue}{30.55}                       & 0.8463                        & 29.21                           & 0.8081          & 28.69          & 0.8624          & 34.00          & 0.9478  \\
			ESWT \cite{ESWT}                  &                             & 578                        & 34.63            & 0.9290            & \color{blue}{30.55}              & 0.8464               & 29.23                  & 0.8088 & 28.70 & 0.8628 & 34.05 & 0.9479 
			\\
			EDT-T \cite{EDT}       &                             & 919                     & \color{blue}{34.73}                     & \color{blue}{0.9299}                    & \color{red}{30.71}                       & \color{blue}{0.8481}                        & \color{blue}{29.29}                           & \color{blue}{0.8103}          & \color{blue}{28.89}          & \color{blue}{0.8674}          & \color{blue}{34.44}          & \color{blue}{0.9498}  
			\\
			EMT(our)       &                          & 678                        & \color{red}{34.80}                  & \color{red}{0.9303}                    & \color{red}{30.71}                       & \color{red}{0.8489}                       & \color{red}{29.33}                          & \color{red}{0.8113}          &\color{red}{29.16}         &\color{red}{0.8716}          & \color{red}{34.65}          & \color{red}{0.9508}  
			\\
			
			\specialrule{0em}{1pt}{0pt}
			\hline
			\specialrule{0em}{2pt}{0pt}
			
			FDIWN \cite{FDIWN}           &  \multirow{9}{*}{$\times4$}                           & 664                       & 32.23                     & 0.8955                     & 28.66                       & 0.7829                        & 27.62                           & 0.7380          & 26.28          & 0.7919          & -              & -  \\
			LBNet \cite{LBNet}           &                             & 742                 & 32.29                     & 0.8960                     & 28.68                       & 0.7832                        & 27.62                           & 0.7382          & 26.27          & 0.7906          & 30.76          & 0.9111  \\
			LAPAR-A \cite{LAPAR}           &                             & 659                  & 32.15                     & 0.8944                     & 28.61                      & 0.7818                        & 27.61                       & 0.7366         & 26.14          & 0.7871          & 30.42         & 0.9074  \\
			NGswin \cite{NGswin}                &                             & 1,019                    & 32.33           & 0.8963            & 28.78              &  0.7859           & 27.66          & 0.7396            & 26.45              &  0.7963                              & 30.80 & 0.9128  \\
			SwinIR \cite{swinir}         &                             & 897                      & 32.44                     & 0.8976                     & 28.77                       & 0.7858                        & 27.69                           & 0.7406          & 26.47          & 0.7980          & 30.92          & 0.9151 \\
			ELAN-light \cite{elan}       &                             & 601                    & 32.43                     & 0.8975                     & 28.78                       & 0.7858                        & 27.69                           & 0.7406          & 26.54          & 0.7982          & 30.92          & 0.9150  \\
			ESWT \cite{ESWT}                &                             & 589                     & 32.46            & 0.8979            & 28.80              & 0.7866               & 27.70            & 0.7410            & 26.56              & 0.8006                              & 30.94 & 0.9136  \\
			EDT-T \cite{EDT}                  &                             & 922                      & \color{blue}{32.53}          &  \color{blue}{0.8991}            &  \color{blue}{28.88}              & \color{blue}0.7882               & \color{blue}27.76                  &  \color{blue}{0.7433} &  \color{blue}{26.71} & \color{blue}0.8051 &  \color{blue}{31.35} &  \color{blue}{0.918}  \\
			EMT(ours)                  &                             & 690                    & \color{red}{32.64}            & \color{red}{0.9003}            & \color{red}{28.97}              & \color{red}{0.7901}               & \color{red}{27.81}                  & \color{red}{0.7441} & \color{red}{26.98} &\color{red}{0.8118} & \color{red}{31.48} & \color{red}{0.9190}  \\
			
			\specialrule{0em}{1pt}{0pt}
			\hline

		\end{tabu}
	}
	
\end{table*}
\section{Experiments}
\label{experiments}
\subsection{Implementation Details}
\label{train-set}
\paragraph{Datasets and Metrics.} 
We use DF2K (DIV2K \cite{DIV2K}, Flickr2K \cite{EDSR}) as the training set,  DIV2K \cite{DIV2K} dataset contains HR images with various scenes and objects, and  Flickr2K \cite{EDSR} dataset contains images with multiple quality levels. The Set5 \cite{set5}, Set14 \cite{set14}, BSD100 \cite{BSD100}, Urban100 \cite{U100}, and Manga109 \cite{M109} datasets are used as the test set to evaluate the performance of our method. The Peak Signal-to-Noise Ratio (PSNR) and Structural Similarity Index Measure (SSIM) are used as evaluation metrics, where the RGB are first converted to YCbCr format, and then the metrics are computed on the Y channel. In addition, we report the network parameters to compare our method with other state-of-the-art (SOTA). The network parameters indicate the model complexity and the amount of computational resources required to train and use the model. 

\paragraph{Training Setting.}
 In proposed method, we set the channel input to 60. In the DFEU, the number of MTBs is set to six. Each MTB consists of six layers (2GTL and 4LTL). The number of SWSA heads is three and the mutual vertical window is ((32, 8), (8, 32)). During the training process, we use data augmentation techniques, including random rotations of $90^\circ$, $180^\circ$, and $270^\circ$, and horizontal flipping. The batch size is set to 64 and the input patch size of LR is $64\times64$. We use the Adam optimizer \cite{Adam} with $\beta_1$ = 0.9 and $\beta_2$ = 0.999 for model optimization. The initial learning rate is set to $5\times10^{-4}$ and decays to $1\times10^{-6}$ using cosine annealing scheduler \cite{cosine}. To train our model, we use the $L_1$ loss function for a total of $1\times10^6$ iterations. The training process is carried out on two NVIDIA V100 32G GPUs using the Pytorch \cite{pytorch} deep learning framework. The proposed training settings and optimization techniques help to ensure efficient and effective model training, leading to SOTA performance on benchmark dataset.

\begin{figure*}[t]
	\centering
	\begin{subfigure}{0.95\linewidth}
		\centering
		\includegraphics[width=1\linewidth]{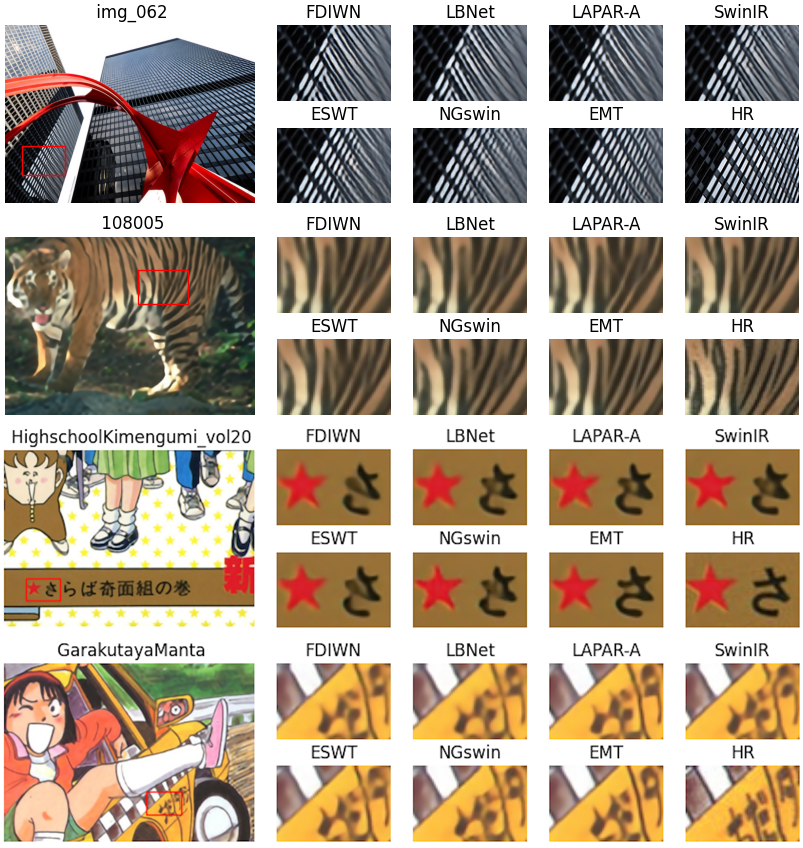}
	\end{subfigure}
	\caption{Qualitative comparison of SOTA methods on the $\times$4 test set, the reconstructed image of EMT is sharper, with fewer artifacts, allowing better recovery of the structure.}
	\label{qualitative_comparison_1}%文中引用该图片代号
\end{figure*}

\subsection{Comparison with SOTA Methods}
\paragraph{Quantitative comparison.} According to the comparison in Table \ref{tab: comparison}, our method outperforms other SOTA, such as SwinIR \cite{swinir}, ELAN-light \cite{elan}, and EDT-T \cite {EDT}, while maintaining the parameters at lightweight scale. Transformer-based methods utilize SA to model the global dependence for the input image, and outperform many CNN-based methods. Later, EDT-T \cite{EDT} used a pre-training strategy to achieve outstanding results, even surpassing SwinIR \cite{swinir} but still lower than our method. Our method is benefited from a combination of PM and SWSA to effectively capture local knowledge and global connectivity, and achieves superior results on various test sets, demonstrating the potential of our method in SR tasks. 

\paragraph{Qualitative comparison.}
We qualitatively compare the SR quality of these different lightweight methods, and the results are shown in Fig. \ref{qualitative_comparison_1}. Given that ELAN-light \cite{elan} and EDT-T \cite{EDT} are not officially provided models, they have been omitted from the comparison. According to the figure, CNN-based methods, such as FDIWN, LBNet, in image\_062 of Urban100 \cite{U100} shows severe artifacts in the construction of the details.  SwinIR \cite{swinir} outperforms CNN-based methods in terms of detail construction, but several artifacts and smoothing problems still appear. Our method further relieves this problem and allows for clearer image recovery. In 108005 image \cite{BSD100}, our method recovers clearer detailed textures than other transformer-based methods due to the enhanced local knowledge interaction. For HighschoolKimengumi-vol20 and GarakutayaManta of Manga109 \cite{M109}, our method is more accurate in font recover, less prone to generate misspellings and therefore more reliable. Overall, the details of SR images in EMT are clearer and more realistic, and the structural information is more obvious due to the enhanced local knowledge aggregation and efficient global interaction.
\begin{figure*}[h] %加*是跨两栏
	\centering
	\begin{subfigure}{0.29\linewidth}
		\includegraphics[width=0.84\linewidth]{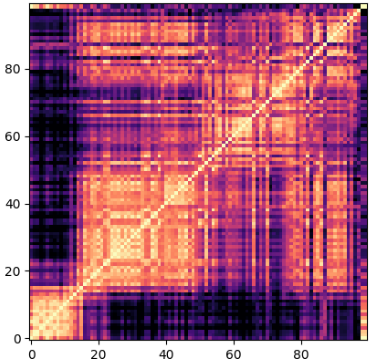}
		\caption{MTB-2GTL}
		\label{2_GTL}%文中引用该图片代号
	\end{subfigure}
	\begin{subfigure}{0.29\linewidth}
		\includegraphics[width=0.845\linewidth]{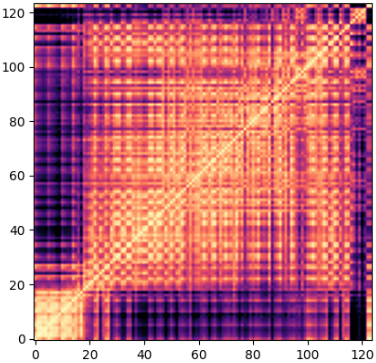}
		\caption{MTB-4GTL}                                                                                                                                                                   
		\label{4_GTL}%文中引用该图片代号
	\end{subfigure}
	\begin{subfigure}{0.29\linewidth}
		\includegraphics[width=1\linewidth]{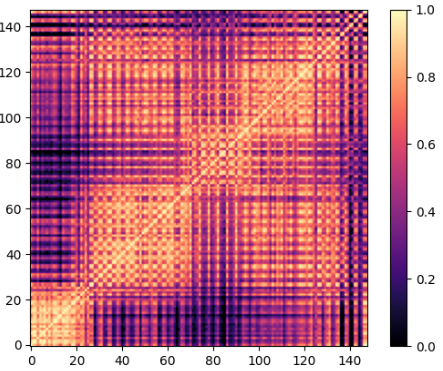}
		\caption{MTB-6GTL}
		\label{6_GTL}%文中引用该图片代号
	\end{subfigure}
	\caption{CKA similarity between MTB layers using different numbers of GTLs, with horizontal and vertical coordinates indicating network depth.}
	\label{CKA_image}
\end{figure*}
\subsection{Ablation studies}
\label{Ablation-studies}
\paragraph{Similarity between Layers.}
Central kernel alignment (CKA) is commonly used to study the representational similarity of hidden layers in networks \cite{EDT,CKA}. We introduced this method to investigate the extraction of features using MTB with different numbers of GTLs. Specifically, 
given $m$ data points, we input the activation $\textbf{X}\in\mathbb{R}^{m \times p_1}$ and $\textbf{Y}\in\mathbb{R}^{m \times p_2}$ of two layers, having $p_1$ and $p_2$ neurons, respectively, as follows:
\begin{equation}
	\rm{CKA}(\textbf{K},\textbf{L})= \frac{\rm{HSIC}(\textbf{K},\textbf{L})}{ \sqrt{\rm{HSIC}(\textbf{K},\textbf{K})\rm{HSIC}(\textbf{L},\textbf{L})}},
\end{equation}
where we use $m \times m$ Gram matrices $\textbf{K}= \textbf{X}\textbf{X}^\top$ and $\textbf{L}= \textbf{Y}\textbf{Y}^\top$ and HSIC is the Hilbert-Schmidt independence criterion \cite{Hilbert-Schmidt}. To facilitate the experiment, we use minibatches of size $n$ = 288, with six layers in each block and a batch size of eight for training strategy, and otherwise the same as above \ref{train-set}. For the fairness of experiments, the position of $TokenMixer$ \cite{mateformer} in LTL is replaced is replaced by $Identity( \cdot )$, and SWSA is used in the $TokenMixer$ position of GTL. MTBs with different numbers of GTLs (2, 4, 6) were used in the experiments and the same training strategy was performed. The results are presented in Fig. \ref{CKA_image}, and the lower left and upper right corners are extracted from SFEU and RECU. Similar yellow squares can be observed on the heat map of the initial and intermediate hidden layers, of which the lighter the colour, the higher the similarity. Comparing MTB-4GTL and -6GTL with the MTB-2GTL, the yellow area is larger and lighter, which may indicate the presence of redundant operations in the network \cite{CKA}. On this basis, we suggest that reducing the number of GTLs to 2 may be a more reasonable choice.

\begin{figure*}[t] %加*是跨两栏
	\centering
	\begin{subfigure}{0.31\linewidth}
		\includegraphics[width=1.02\linewidth]{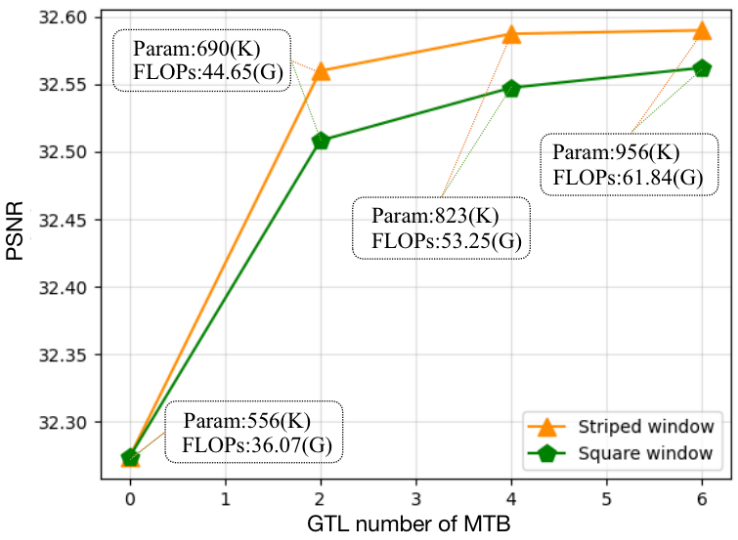}
		\caption{Set5 \cite{set5}}                                                                                                                                                                   
	\end{subfigure}
	\begin{subfigure}{0.31\linewidth}
		\includegraphics[width=1.01\linewidth]{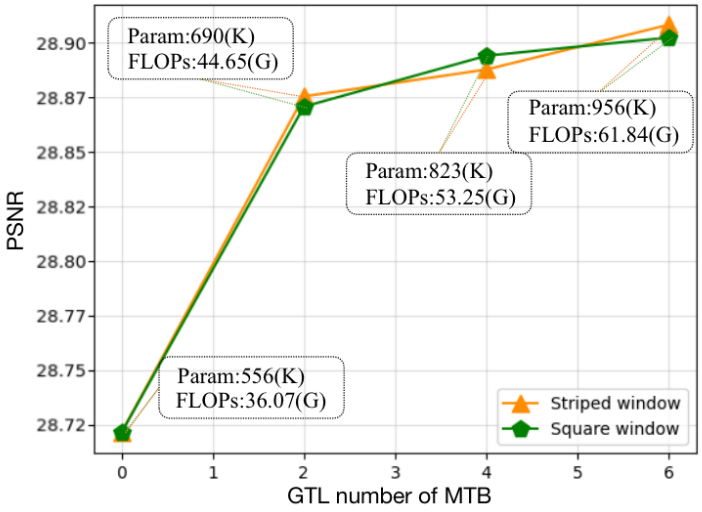}
		\caption{Set14 \cite{set14}}
	\end{subfigure}
	\begin{subfigure}{0.31\linewidth}
		\includegraphics[width=1.02\linewidth]{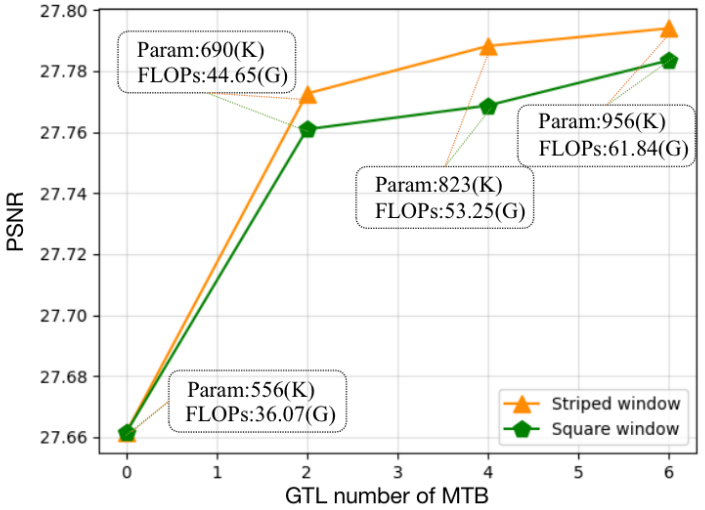}
		\caption{BSD100 \cite{BSD100}}
	\end{subfigure}
	\caption{Comparison of PNSR and parameters using different numbers of GTL in MTB structure on the $\times$4 test set .}
	\label{param_psnr}
\end{figure*}

\paragraph{Effect of MTB structure on Performance.}
GTL is replaced by LTL in the MTB structure to enhance the capture of local knowledge and reduce the parameters and FLOPs, and different levels of replacement result in different performance. In this subsection, we verify the effectiveness of the proposed MTB structure. In GTL, square ($16 \times 16$) and striped windows SA are used, while maintaining the same complexity. As shown in Fig. \ref{param_psnr}, MTB using a larger number of GTLs gradually improves PSNR on the
Set5 \cite{set5}, Set14 \cite{set14}, and BSD100 \cite{BSD100} $\times$4 test set. However, regardless of the type of SA window, we observed a decreasing trend of improvement in PSNR with increasing GTLs, while parameters and FLOPs still increased proportionally. The slope of 0 to 2GTLs in MTB is the largest, and we believe that MTB-2GTL is a cost-effective choice. The striped self-attention window utilizes image anisotropy features and performs better in the test set than the traditional square window. In addition, we further validated the MTB architecture by quantitatively comparing MTB-4GTL and -6GTL with MTB-2GTL, while keeping the same parameter levels. As shown in the Table \ref{tab: as_et2}, MTB-2GTL outperformed the number of other GTLs on most of the test sets. Due to the lack of interaction between the SAs of MTB-1GTL, -3GTL and -5GTL, which may lead to poor results, they are omitted in the display.
\begin{table}[h]
	\begin{minipage}{0.52\linewidth}
		\centering
		\footnotesize
		\hspace{1mm}
		\caption{The PSNR is tested on a set of $\times 4$ image SR, using different numbers of GTLs on the MTB structure while controlling the parameters to be approximately equal. The BEST results are \textbf{highlighted}.}
		\vspace{2mm}
		\resizebox{0.98\linewidth}{!}{
			\begin{tabu}{l|cc|ccccc}
				\toprule
				\multirow{2}{1cm}{Model}  &Params  &FLOPs   & Set5          & Set14          & Urban100    & BSD100     & Manga109  \\
				&  (K) &(G)& (PSNR)        & (PSNR)         & (PSNR)         & (PSNR)  & (PSNR)  \\
				\midrule
				MTB - 0GTL   &733 &47.5 & 32.328 & 28.719 & 26.387 &  27.688& 30.942 \\          
				\rowcolor{Gray} 
				MTB - 2GTL    &690&44.6 & \textbf{32.559} & \textbf{28.875} & 26.741  &  \textbf{27.772}&  \textbf{31.296} \\
				MTB - 4GTL    &690&44.6 & 32.546 & 28.869 &  \textbf{26.760} & 27.757 & 31.277 \\ 
				MTB - 6GTL   &645 &41.8 & 32.491 & 28.864 & 26.754 & 27.757& 31.264 \\
				\bottomrule
			\end{tabu} 
		}
		\captionsetup{font=footnotesize}

		\label{tab: as_et2}    
	\end{minipage}
	\quad
	\begin{minipage}{0.47\linewidth}
		\centering
		\footnotesize
		
		\hspace{0mm}
		\caption{Test results on Manga109 \cite{M109} of $\times 4$ image SR on which module is used in the location of the $TokenMixer$ in LTL. The BEST results are \textbf{highlighted}.	}
		\vspace{3.5mm}
		\vspace{-1.5mm}
		\resizebox{1.0\linewidth}{!}{
			\begin{tabu}{l|cc|cc|c}  %tabullar 竖线不一样
				\toprule
				\multirow{2}{1cm}{Model}     & \multirow{1}{1cm}{Params  } & \multirow{1}{1cm}{FLOPs  } &\multirow{2}{2cm}{$Pixel Mixer(\cdot)$}       &  \multirow{2}{2cm}{ $Identity(\cdot)$}        & \multirow{2}{2cm}{Manga109}         \\
				
				& (K)  &(G)   &        &          &             \\
				
				\midrule
				\rowcolor{Gray}  
				MTB - 4GTL   &  690& 44.6& \checkmark &  & \textbf{31.328}  \\
				
				MTB - 4GTL   &  690 &44.6&  & \checkmark  & 31.277  \\
				\midrule
				\rowcolor{Gray}   
				MTB - 2GTL   & 690  &44.6& \checkmark  &  & \textbf{31.329}  \\
				
				MTB - 2GTL   & 690 &44.6&  & \checkmark  & 31.296 \\
				
				\bottomrule
			\end{tabu} 
			
			\captionsetup{font=footnotesize}    
			
			% All model are trained with 640$\times$640 resolution.
		}
		\label{tab: as_et3}
	\end{minipage}
	
\end{table}

\begin{figure*}[h]
	\centering
	\begin{subfigure}{0.95\linewidth}
		\centering
		\includegraphics[width=1\linewidth]{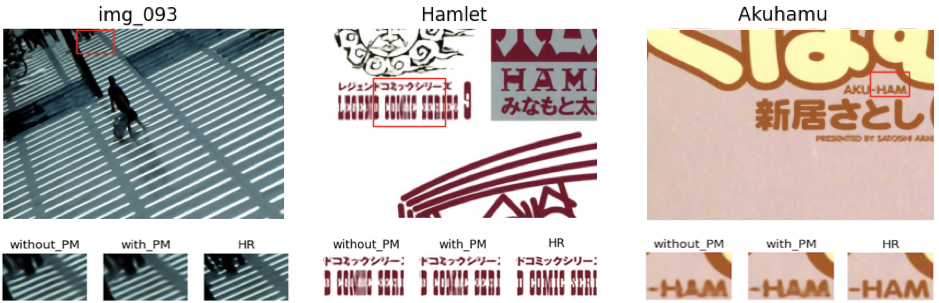}
	\end{subfigure}
	\caption{Qualitative analysis results of MTB with and without PM in Urban100 \cite{U100} and Manga109 \cite{M109} ($\times4$) test sets.}
	\label{psnr_pdim}%文中引用该图片代号
\end{figure*}

\begin{figure*}[h] %加*是跨两栏
	\centering
	\begin{subfigure}{0.9\linewidth}
		\centering
		\includegraphics[width=0.9\linewidth]{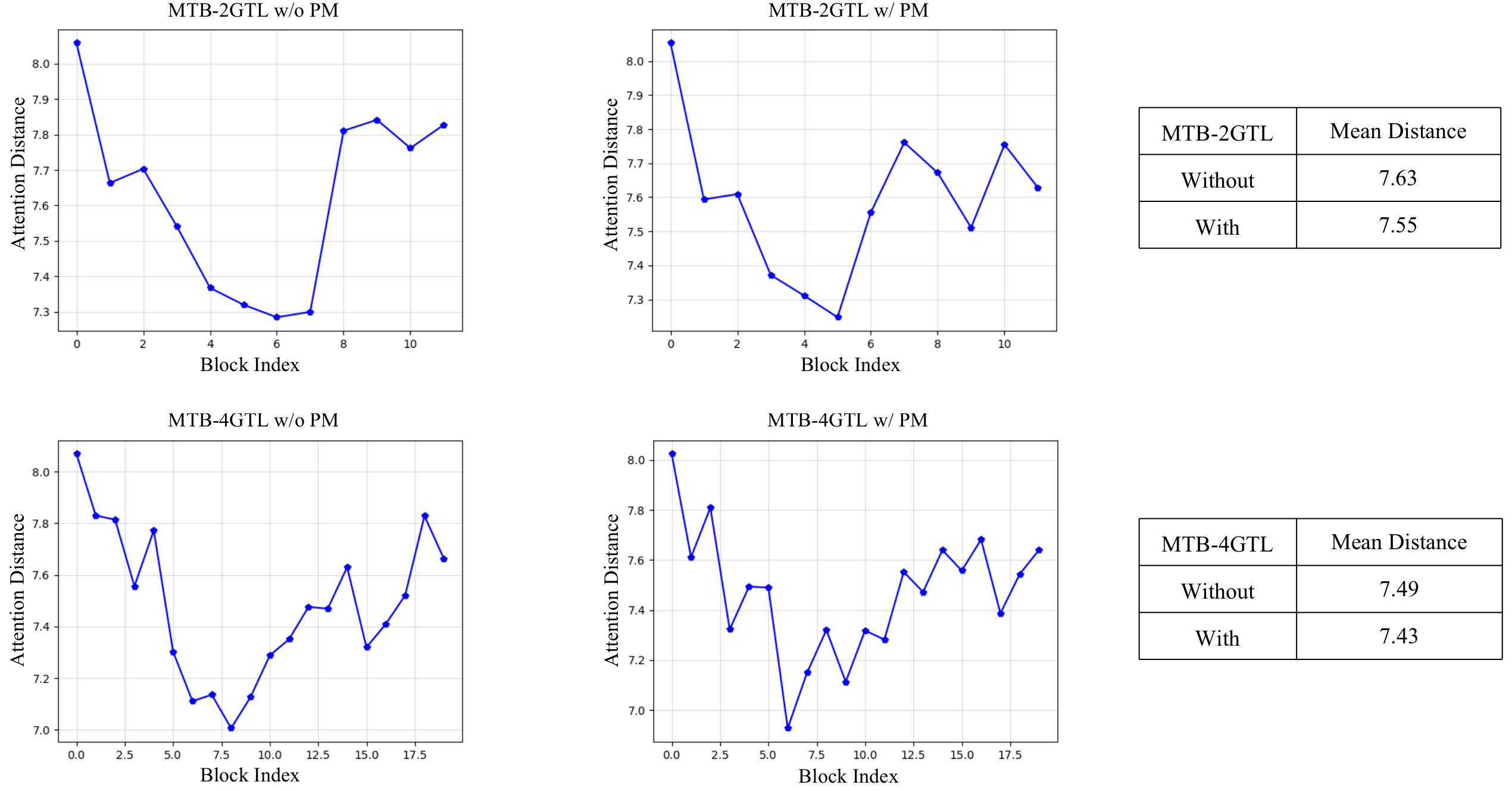}
		\label{mad-4SA-PM-ID}%文中引用该图片代号
	\end{subfigure}
	\caption{The distance of attention head for the MTB with and without PM.}
	\label{mad_image}
\end{figure*}

\paragraph{Effectiveness of the Pixel Mixer.}
We propose PM to enhance locality mechanisms in the architecture by mixing pixels without adding parameters and FLOPs. To verify the effectiveness of PM, we set up four groups of experiments for quantitative and qualitative analysis, i.e., with and without PM in MTB-2GTL and MTB-4GTL, as shown in Table \ref{tab: as_et3} and Fig. \ref{psnr_pdim}. The results show that the network with PM improves the PSNR of test set and recovers clearer and more realistic image details and textures while maintaining the same parameter levels. In addition, the ability of the network to utilize local knowledge can be reflected by observing the change in Mean Attention Distance (MAD) \cite{EDT}. MAD is obtained by averaging the distance between the query pixel and all other pixels, weighted by the attention weights. The points represent the attention distance with lower indexes typically indicating increased use of local knowledge in the network. We further carry out MAD experiments on the DIV2K \cite{DIV2K} validation set. As shown in Fig. \ref{mad_image}, PM brought more locality in several regions, especially in the higher network layers, and the overall mean attention distance decreased. The introduction of PM has no additional network parameters and enhances the capability of the network to aggregate local knowledge.

\section{Conclusion}
This study proposes an Efficint Mixed Transformer (EMT) for SISR, which consists of three units: shallow feature extraction, deep feature extraction, and reconstruction. The deep feature extraction unit uses Mixed Transformer Block (MTB) with the mixture of global transformer layer (GTL) and local transformer layer (LTL) in each block. LTL mainly consists of a Pixel Mixer (PM)  and a multi-layer perceptron. PM enhances the locality mechanism of the network with channel separation and pixel shifting operations without additional complexity. Striped window for self-attention (SWSA) in GTL utilizes the anisotropy of images to obtaining a more effective global dependency modelling. The experimental results show that EMT achieves more outstanding performance in LSR than previous SOTA methods and a better balance between performance and complexity. In the future, we attempt to reduce the complexity of self-attention and consider SISR deployment in mobile and embedded devices.
{\small
	\bibliographystyle{plain}
	\bibliography{neurips.bib}

\begin{thebibliography}{46}
\providecommand{\natexlab}[1]{#1}
\providecommand{\url}[1]{\texttt{#1}}
\expandafter\ifx\csname urlstyle\endcsname\relax
  \providecommand{\doi}[1]{doi: #1}\else
  \providecommand{\doi}{doi: \begingroup \urlstyle{rm}\Url}\fi

\bibitem[Agustsson and Timofte(2017)]{DIV2K}
Eirikur Agustsson and Radu Timofte.
\newblock Ntire 2017 challenge on single image super-resolution: Dataset and
  study.
\newblock In \emph{Proceedings of the IEEE conference on computer vision and
  pattern recognition workshops}, pages 126--135, 2017.

\bibitem[Bevilacqua et~al.(2012)Bevilacqua, Roumy, Guillemot, and Morel]{set5}
M.~Bevilacqua, A.~Roumy, C.~Guillemot, and A.~Morel.
\newblock Low-complexity single image super-resolution based on nonnegative
  neighbor embedding.
\newblock In \emph{British Machine Vision Conference}, 2012.

\bibitem[Chen et~al.(2021)Chen, Gu, and Zhang]{DeepSR3}
Haoyu Chen, Jinjin Gu, and Zhi Zhang.
\newblock Attention in attention network for image super-resolution.
\newblock \emph{arXiv preprint arXiv:2104.09497}, 2021.

\bibitem[Choi et~al.(2022)Choi, Lee, and Yang]{NGswin}
Haram Choi, Jeongmin Lee, and Jihoon Yang.
\newblock N-gram in swin transformers for efficient lightweight image
  super-resolution.
\newblock \emph{arXiv preprint arXiv:2211.11436}, 2022.

\bibitem[Deng et~al.(2009)Deng, Dong, Socher, Li, Li, and Fei-Fei]{imagenet}
Jia Deng, Wei Dong, Richard Socher, Li-Jia Li, Kai Li, and Li~Fei-Fei.
\newblock Imagenet: A large-scale hierarchical image database.
\newblock In \emph{2009 IEEE conference on computer vision and pattern
  recognition}, pages 248--255. Ieee, 2009.

\bibitem[Dong et~al.(2014)Dong, Loy, He, and Tang]{SRCNN}
Chao Dong, Chen~Change Loy, Kaiming He, and Xiaoou Tang.
\newblock Learning a deep convolutional network for image super-resolution.
\newblock In \emph{European conference on computer vision}, pages 184--199.
  Springer, 2014.

\bibitem[Dong et~al.(2015)Dong, Loy, He, and Tang]{SRusedconv1}
Chao Dong, Chen~Change Loy, Kaiming He, and Xiaoou Tang.
\newblock Image super-resolution using deep convolutional networks.
\newblock \emph{IEEE transactions on pattern analysis and machine
  intelligence}, 38\penalty0 (2):\penalty0 295--307, 2015.

\bibitem[Gao et~al.(2022{\natexlab{a}})Gao, Li, Li, Wu, Lu, and Yu]{FDIWN}
Guangwei Gao, Wenjie Li, Juncheng Li, Fei Wu, Huimin Lu, and Yi~Yu.
\newblock Feature distillation interaction weighting network for lightweight
  image super-resolution.
\newblock In \emph{AAAI Conference on Artificial Intelligence (AAAI)},
  2022{\natexlab{a}}.

\bibitem[Gao et~al.(2022{\natexlab{b}})Gao, Wang, Li, Li, Yu, and Zeng]{LBNet}
Guangwei Gao, Zhengxue Wang, Juncheng Li, Wenjie Li, Yi~Yu, and Tieyong Zeng.
\newblock Lightweight bimodal network for single-image super-resolution via
  symmetric cnn and recursive transformer.
\newblock In \emph{International Joint Conference on Artificial Intelligence
  (IJCAI)}, 2022{\natexlab{b}}.

\bibitem[Gretton et~al.(2007)Gretton, Fukumizu, Teo, Le, and
  Smola]{Hilbert-Schmidt}
A.~Gretton, K.~Fukumizu, C.~H. Teo, S.~Le, and A.~J. Smola.
\newblock A kernel statistical test of independence.
\newblock In \emph{Advances in Neural Information Processing Systems 20,
  Proceedings of the Twenty-First Annual Conference on Neural Information
  Processing Systems, Vancouver, British Columbia, Canada, December 3-6, 2007},
  2007.

\bibitem[Han et~al.(2021)Han, Fan, Dai, Sun, and Wang]{DWconv_SA}
Q.~Han, Z.~Fan, Q.~Dai, L.~Sun, and J.~Wang.
\newblock Demystifying local vision transformer: Sparse connectivity, weight
  sharing, and dynamic weight, 2021.

\bibitem[Haris et~al.(2018)Haris, Shakhnarovich, and Ukita]{SRusedconv3}
Muhammad Haris, Gregory Shakhnarovich, and Norimichi Ukita.
\newblock Deep back-projection networks for super-resolution.
\newblock In \emph{Proceedings of the IEEE conference on computer vision and
  pattern recognition}, pages 1664--1673, 2018.

\bibitem[Huang et~al.(2017)Huang, Liu, Van Der~Maaten, and Weinberger]{PLoss1}
Gao Huang, Zhuang Liu, Laurens Van Der~Maaten, and Kilian~Q Weinberger.
\newblock Densely connected convolutional networks.
\newblock In \emph{Proceedings of the IEEE conference on computer vision and
  pattern recognition}, pages 4700--4708, 2017.

\bibitem[Huang et~al.(2015)Huang, Singh, and Ahuja]{U100}
Jia-Bin Huang, Abhishek Singh, and Narendra Ahuja.
\newblock Single image super-resolution from transformed self-exemplars.
\newblock In \emph{Proceedings of the IEEE conference on computer vision and
  pattern recognition}, pages 5197--5206, 2015.

\bibitem[Kim et~al.(2016)Kim, Lee, and Lee]{VDSR}
Jiwon Kim, Jung~Kwon Lee, and Kyoung~Mu Lee.
\newblock Accurate image super-resolution using very deep convolutional
  networks.
\newblock In \emph{Proceedings of the IEEE conference on computer vision and
  pattern recognition}, pages 1646--1654, 2016.

\bibitem[Kingma and Ba(2014)]{Adam}
Diederik~P Kingma and Jimmy Ba.
\newblock Adam: A method for stochastic optimization.
\newblock \emph{arXiv preprint arXiv:1412.6980}, 2014.

\bibitem[Lai et~al.(2017{\natexlab{a}})Lai, Huang, Ahuja, and Yang]{DeepSR1}
Wei-Sheng Lai, Jia-Bin Huang, Narendra Ahuja, and Ming-Hsuan Yang.
\newblock Deep laplacian pyramid networks for fast and accurate
  super-resolution.
\newblock In \emph{Proceedings of the IEEE conference on computer vision and
  pattern recognition}, pages 624--632, 2017{\natexlab{a}}.

\bibitem[Lai et~al.(2017{\natexlab{b}})Lai, Huang, Ahuja, and
  Yang]{SRusedconv4}
Wei-Sheng Lai, Jia-Bin Huang, Narendra Ahuja, and Ming-Hsuan Yang.
\newblock Deep laplacian pyramid networks for fast and accurate
  super-resolution.
\newblock In \emph{Proceedings of the IEEE conference on computer vision and
  pattern recognition}, pages 624--632, 2017{\natexlab{b}}.

\bibitem[LeCun et~al.(2015)LeCun, Bengio, and Hinton]{DeepSR0}
Yann LeCun, Yoshua Bengio, and Geoffrey Hinton.
\newblock Deep learning.
\newblock \emph{nature}, 521\penalty0 (7553):\penalty0 436--444, 2015.

\bibitem[Ledig et~al.(2017)Ledig, Theis, Husz{\'a}r, Caballero, Cunningham,
  Acosta, Aitken, Tejani, Totz, Wang, et~al.]{SRphoto}
Christian Ledig, Lucas Theis, Ferenc Husz{\'a}r, Jose Caballero, Andrew
  Cunningham, Alejandro Acosta, Andrew Aitken, Alykhan Tejani, Johannes Totz,
  Zehan Wang, et~al.
\newblock Photo-realistic single image super-resolution using a generative
  adversarial network.
\newblock In \emph{Proceedings of the IEEE conference on computer vision and
  pattern recognition}, pages 4681--4690, 2017.

\bibitem[Li et~al.(2021{\natexlab{a}})Li, Lu, Lu, Zhang, and Jia]{EDT}
W.~Li, X.~Lu, J.~Lu, X.~Zhang, and J.~Jia.
\newblock On efficient transformer and image pre-training for low-level vision.
\newblock \emph{arXiv e-prints}, 2021{\natexlab{a}}.

\bibitem[Li et~al.(2020)Li, Zhou, Qi, Jiang, Lu, and Jia]{LAPAR}
Wenbo Li, Kun Zhou, Lu~Qi, Nianjuan Jiang, Jiangbo Lu, and Jiaya Jia.
\newblock Lapar: Linearly-assembled pixel-adaptive regression network for
  single image super-resolution and beyond.
\newblock In \emph{Advances in Neural Information Processing Systems
  (NeurIPS)}, 2020.

\bibitem[Li et~al.(2021{\natexlab{b}})Li, Zhang, Cao, Timofte, and
  Van~Gool]{Local_SA1}
Yawei Li, Kai Zhang, Jiezhang Cao, Radu Timofte, and Luc Van~Gool.
\newblock Localvit: Bringing locality to vision transformers.
\newblock \emph{arXiv preprint arXiv:2104.05707}, 2021{\natexlab{b}}.

\bibitem[Li et~al.(2023)Li, Fan, Xiang, Demandolx, Ranjan, Timofte, and
  Van~Gool]{IR-stripes-SA}
Yawei Li, Yuchen Fan, Xiaoyu Xiang, Denis Demandolx, Rakesh Ranjan, Radu
  Timofte, and Luc Van~Gool.
\newblock Efficient and explicit modelling of image hierarchies for image
  restoration.
\newblock \emph{arXiv preprint arXiv:2303.00748}, 2023.

\bibitem[Liang et~al.(2021)Liang, Cao, Sun, Zhang, Van~Gool, and
  Timofte]{swinir}
Jingyun Liang, Jiezhang Cao, Guolei Sun, Kai Zhang, Luc Van~Gool, and Radu
  Timofte.
\newblock Swinir: Image restoration using swin transformer.
\newblock In \emph{Proceedings of the IEEE/CVF International Conference on
  Computer Vision}, pages 1833--1844, 2021.

\bibitem[Lim et~al.(2017)Lim, Son, Kim, Nah, and Mu~Lee]{EDSR}
Bee Lim, Sanghyun Son, Heewon Kim, Seungjun Nah, and Kyoung Mu~Lee.
\newblock Enhanced deep residual networks for single image super-resolution.
\newblock In \emph{Proceedings of the IEEE conference on computer vision and
  pattern recognition workshops}, pages 136--144, 2017.

\bibitem[Liu et~al.(2021)Liu, Lin, Cao, Hu, Wei, Zhang, Lin, and
  Guo]{swintrans}
Ze~Liu, Yutong Lin, Yue Cao, Han Hu, Yixuan Wei, Zheng Zhang, Stephen Lin, and
  Baining Guo.
\newblock Swin transformer: Hierarchical vision transformer using shifted
  windows.
\newblock In \emph{Proceedings of the IEEE/CVF International Conference on
  Computer Vision}, pages 10012--10022, 2021.

\bibitem[Loshchilov and Hutter(2016)]{cosine}
I.~Loshchilov and F.~Hutter.
\newblock Sgdr: Stochastic gradient descent with restarts.
\newblock 2016.

\bibitem[Lu et~al.(2022)Lu, Li, Liu, Huang, Zhang, and Zeng]{ESRT}
Zhisheng Lu, Juncheng Li, Hong Liu, Chaoyan Huang, Linlin Zhang, and Tieyong
  Zeng.
\newblock Transformer for single image super-resolution.
\newblock In \emph{Proceedings of the IEEE/CVF Conference on Computer Vision
  and Pattern Recognition}, pages 457--466, 2022.

\bibitem[Martin et~al.(2001)Martin, Fowlkes, Tal, and Malik]{BSD100}
David Martin, Charless Fowlkes, Doron Tal, and Jitendra Malik.
\newblock A database of human segmented natural images and its application to
  evaluating segmentation algorithms and measuring ecological statistics.
\newblock In \emph{Proceedings Eighth IEEE International Conference on Computer
  Vision. ICCV 2001}, volume~2, pages 416--423. IEEE, 2001.

\bibitem[Matsui et~al.(2017)Matsui, Ito, Aramaki, Fujimoto, Ogawa, Yamasaki,
  and Aizawa]{M109}
Yusuke Matsui, Kota Ito, Yuji Aramaki, Azuma Fujimoto, Toru Ogawa, Toshihiko
  Yamasaki, and Kiyoharu Aizawa.
\newblock Sketch-based manga retrieval using manga109 dataset.
\newblock \emph{Multimedia Tools and Applications}, 76:\penalty0 21811--21838,
  2017.

\bibitem[Nguyen et~al.(2020)Nguyen, Raghu, and Kornblith]{CKA}
Thao Nguyen, Maithra Raghu, and Simon Kornblith.
\newblock Do wide and deep networks learn the same things? uncovering how
  neural network representations vary with width and depth.
\newblock \emph{arXiv preprint arXiv:2010.15327}, 2020.

\bibitem[Paszke et~al.(2019)Paszke, Gross, Massa, Lerer, Bradbury, Chanan,
  Killeen, Lin, Gimelshein, Antiga, et~al.]{pytorch}
Adam Paszke, Sam Gross, Francisco Massa, Adam Lerer, James Bradbury, Gregory
  Chanan, Trevor Killeen, Zeming Lin, Natalia Gimelshein, Luca Antiga, et~al.
\newblock Pytorch: An imperative style, high-performance deep learning library.
\newblock \emph{Advances in neural information processing systems}, 32, 2019.

\bibitem[Sajjadi et~al.(2017{\natexlab{a}})Sajjadi, Scholkopf, and
  Hirsch]{DeepSR2}
Mehdi~SM Sajjadi, Bernhard Scholkopf, and Michael Hirsch.
\newblock Enhancenet: Single image super-resolution through automated texture
  synthesis.
\newblock In \emph{Proceedings of the IEEE international conference on computer
  vision}, pages 4491--4500, 2017{\natexlab{a}}.

\bibitem[Sajjadi et~al.(2017{\natexlab{b}})Sajjadi, Scholkopf, and
  Hirsch]{PLoss2}
Mehdi~SM Sajjadi, Bernhard Scholkopf, and Michael Hirsch.
\newblock Enhancenet: Single image super-resolution through automated texture
  synthesis.
\newblock In \emph{Proceedings of the IEEE international conference on computer
  vision}, pages 4491--4500, 2017{\natexlab{b}}.

\bibitem[Shi et~al.(2023)Shi, Li, Liu, Liu, Zhang, Zhu, Zheng, and Weng]{ESWT}
Jinpeng Shi, Hui Li, Tianle Liu, Yulong Liu, Mingjian Zhang, Jinchen Zhu, Ling
  Zheng, and Shizhuang Weng.
\newblock Image super-resolution using efficient striped window transformer.
\newblock \emph{arXiv preprint arXiv:2301.09869}, 2023.

\bibitem[Tai et~al.(2017)Tai, Yang, Liu, and Xu]{MemNet}
Ying Tai, Jian Yang, Xiaoming Liu, and Chunyan Xu.
\newblock Memnet: A persistent memory network for image restoration.
\newblock In \emph{Proceedings of the IEEE international conference on computer
  vision}, pages 4539--4547, 2017.

\bibitem[Vaswani et~al.(2017)Vaswani, Shazeer, Parmar, Uszkoreit, Jones, Gomez,
  Kaiser, and Polosukhin]{transformer}
Ashish Vaswani, Noam Shazeer, Niki Parmar, Jakob Uszkoreit, Llion Jones,
  Aidan~N Gomez, {\L}ukasz Kaiser, and Illia Polosukhin.
\newblock Attention is all you need.
\newblock \emph{Advances in neural information processing systems}, 30, 2017.

\bibitem[Wang et~al.(2015)Wang, Liu, Yang, Han, and Huang]{DeepSR4}
Zhaowen Wang, Ding Liu, Jianchao Yang, Wei Han, and Thomas Huang.
\newblock Deep networks for image super-resolution with sparse prior.
\newblock In \emph{Proceedings of the IEEE international conference on computer
  vision}, pages 370--378, 2015.

\bibitem[Wu et~al.(2017)Wu, Wan, Yue, Jin, Zhao, Golmant, Gholaminejad,
  Gonzalez, and Keutzer]{Shiftc}
B.~Wu, A.~Wan, X.~Yue, P.~Jin, S.~Zhao, N.~Golmant, A.~Gholaminejad,
  J.~Gonzalez, and K.~Keutzer.
\newblock Shift: A zero flop, zero parameter alternative to spatial
  convolutions.
\newblock 2017.

\bibitem[Wu et~al.(2021)Wu, Xiao, Codella, Liu, Dai, Yuan, and
  Zhang]{Local_SA2}
Haiping Wu, Bin Xiao, Noel Codella, Mengchen Liu, Xiyang Dai, Lu~Yuan, and Lei
  Zhang.
\newblock Cvt: Introducing convolutions to vision transformers.
\newblock In \emph{Proceedings of the IEEE/CVF International Conference on
  Computer Vision}, pages 22--31, 2021.

\bibitem[Yu et~al.(2022)Yu, Luo, Zhou, Si, Zhou, Wang, Feng, and
  Yan]{mateformer}
Weihao Yu, Mi~Luo, Pan Zhou, Chenyang Si, Yichen Zhou, Xinchao Wang, Jiashi
  Feng, and Shuicheng Yan.
\newblock Metaformer is actually what you need for vision.
\newblock In \emph{Proceedings of the IEEE/CVF Conference on Computer Vision
  and Pattern Recognition}, pages 10819--10829, 2022.

\bibitem[Zeyde et~al.(2012)Zeyde, Elad, and Protter]{set14}
Roman Zeyde, Michael Elad, and Matan Protter.
\newblock On single image scale-up using sparse-representations.
\newblock In \emph{Curves and Surfaces: 7th International Conference, Avignon,
  France, June 24-30, 2010, Revised Selected Papers 7}, pages 711--730.
  Springer, 2012.

\bibitem[Zhang et~al.(2018{\natexlab{a}})Zhang, Zuo, and Zhang]{SRusedconv2}
Kai Zhang, Wangmeng Zuo, and Lei Zhang.
\newblock Learning a single convolutional super-resolution network for multiple
  degradations.
\newblock In \emph{Proceedings of the IEEE conference on computer vision and
  pattern recognition}, pages 3262--3271, 2018{\natexlab{a}}.

\bibitem[Zhang et~al.(2022)Zhang, Zeng, Guo, and Zhang]{elan}
Xindong Zhang, Hui Zeng, Shi Guo, and Lei Zhang.
\newblock Efficient long-range attention network for image super-resolution.
\newblock \emph{arXiv preprint arXiv:2203.06697}, 2022.

\bibitem[Zhang et~al.(2018{\natexlab{b}})Zhang, Tian, Kong, Zhong, and Fu]{RDN}
Yulun Zhang, Yapeng Tian, Yu~Kong, Bineng Zhong, and Yun Fu.
\newblock Residual dense network for image super-resolution.
\newblock In \emph{Proceedings of the IEEE conference on computer vision and
  pattern recognition}, pages 2472--2481, 2018{\natexlab{b}}.

\end{thebibliography}
}

\newpage
\appendix
%\input{texts4neurips/supp}
%\bibliographystyle{plainnat}
%\bibliography{ref,unsrtnat}
\bibliographystyle{unsrtnat}

\end{document}